\documentclass[a4paper,twoside]{article}

\usepackage{epsfig}
\usepackage{graphicx}
\usepackage{subfigure}
\usepackage{algorithmic,algorithm}
\usepackage{calc}
\usepackage{url}
\usepackage{cite}
\usepackage{amssymb}
\usepackage{amstext}
\usepackage{amsmath}
\usepackage{amsthm}
\usepackage{multicol}
\usepackage{pslatex}
\usepackage{apalike}
\usepackage{SCITEPRESS}
\usepackage[small]{caption}
\begin{document}
\title{Towards an Automated Image De-fencing Algorithm Using Sparsity} %\subtitle{Preparation of Camera-Ready Contributions to SCITEPRESS Proceedings} }

%\author{\authorname{First Author Name\sup{1}, Second Author Name\sup{1} and Third Author Name\sup{2}}
%\affiliation{\sup{1}Institute of Problem Solving, XYZ University, My Street, MyTown, MyCountry}
%\affiliation{\sup{2}Department of Computing, Main University, MySecondTown, MyCountry}
%\email{\{f\_author, s\_author\}@ips.xyz.edu, t\_author@dc.mu.edu}
%}

\author{\authorname{Sankaraganesh Jonna\sup{\dagger}, Krishna K. Nakka\sup{\star} and Rajiv R. Sahay\sup{\dagger,\star}}
\affiliation{Computational Vision Lab} 
\affiliation{\sup{\dagger}School of Information Technology,\sup{\star}Department of Electrical Engineering}
%\affiliation{\sup{2}Department of Electrical Engineering}
\affiliation{Indian Institute of Technology Kharagpur, India}
\email{\{sankar9.iitkgp, krishkanth.92, sahayiitm\}@gmail.com}
}

\keywords{Fence detection, De-fencing, HOG, split Bregman.}
\abstract{Conventional approaches to image de-fencing suffer from non-robust fence detection and are limited to processing images of static scenes. In this position paper, we propose an automatic de-fencing algorithm for images of dynamic scenes. We divide the problem of image de-fencing into the tasks of automated fence detection, motion estimation and fusion of data from multiple frames of a captured video of the dynamic scene. Fences are detected automatically using two approaches, namely, employing Gabor filter and a machine learning method. We cast the fence removal problem in an optimization framework, by modeling the formation of the degraded observations. The inverse problem is solved using split Bregman technique assuming total variation of the de-fenced image as the regularization constraint.}
\onecolumn \maketitle \normalsize \vfill
\section{\uppercase{Introduction}}
\label{sec:introduction}
Images containing fences occur in several situations such as photographing statues in museums, animals in a zoo etc. Image de-fencing involves the removal of fences or occlusions in images. De-fencing a single photo is striclty an image inpainting problem that involves using data in the regions neighbouring the fence pixels in the frame for filling-in occlusions. The works of \cite{Bertalmio,Criminisi_tip,James,Xu,Konstantinos} addressed the image inpainting problem wherein the portion of the image to be inpainted is specified by a mask manually. As shown in Fig. 1(a), in the image de-fencing problem it is difficult to manually mark all fence pixels since they are numerous and cover the entire image. Image inpainting does not yield satisfactory results when the image contains fine textured regions that have to be filled-in. However, using a video panned across a fenced scene can lead to better results due to availability of additional information in the adjacent frames. Image de-fencing using a captured video involves multiple steps such as fence detection, motion estimation and information fusion. Our focus in this position paper is to propose an automatic fence removal system for images of dynamic scenes.
\begin{figure}[t]
\centering     %%% not \center
\subfigure[]{\label{fig:a}\includegraphics[width=24mm]{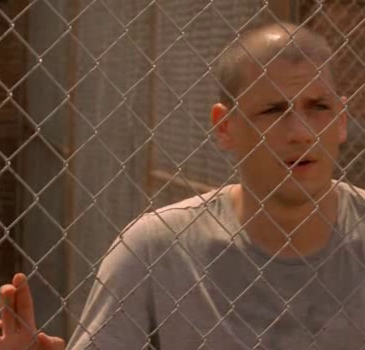}}
\subfigure[]{\label{fig:b}\includegraphics[width=24mm]{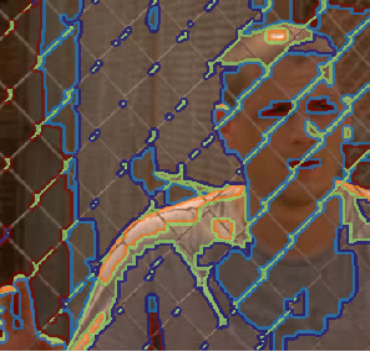}}
\subfigure[]{\label{fig:c}\includegraphics[width=24mm]{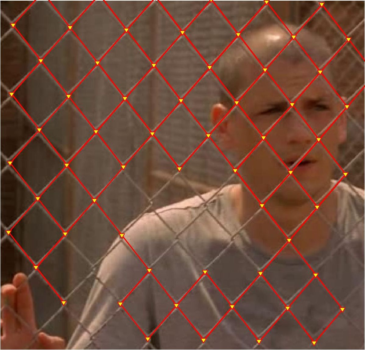}}
%S\subfigure[]{\label{fig:e}\includegraphics[width=40mm]{yanxi_Error_plot}}
\caption{Fence detection: (a) 1st frame from the video. (b) Segmentation result using graph cut \cite{Bagon2006}. (c) Output of the lattice detection algorithm \cite{Minwoo}.}
\end{figure}
 As discussed in \cite{Yanxi,Minwo,Vrushali}, automated fence detection is the first major task in de-fencing. We propose two methods for automated fence detection in this position paper. Firstly, we take advantage of the strongly directional nature of fence occlusions and use a Gabor filter for fence detection. Secondly, using a supervised learning approach, we train an SVM classifier to detect fence pixels automatically. A block diagram of the proposed automatic image de-fencing system is shown in Fig. 2. It involves three major components. Firstly, we need to design an automatic fence detection scheme that should be able to detect fences/occlusions in any complex scene. Secondly, relative motion between the frames has to be estimated. Lastly, we need an algorithm to fuse the information from adjacent frames and produce a de-fenced image. 

\begin{figure*}%name of the figure for reference.....
\centering
\subfigure[]{\includegraphics[width=160mm]{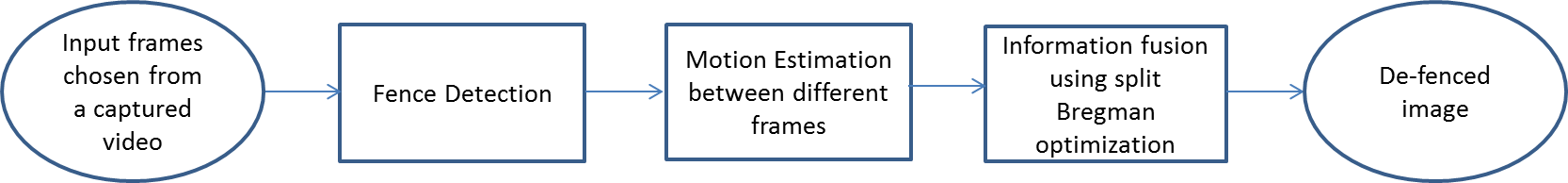}}
\caption{Workflow of proposed automatic image de-fencing algorithm.}
\end{figure*}
Since our goal in this position paper is to automate the above three steps, initially, we propose an automated approach to fence detection in images of dynamic scenes. Next, we estimate the motion between the frames chosen from the video using the optical flow algorithm \cite{Thomas}. Lastly, we formulate an optimization framework for estimating the de-fenced image by solving the corresponding inverse problem. Since natural images are sparse, we use the split Bregman algorithm for optimization with the total variation (TV) of the de-fenced image as the regularization constraint \cite{Tom}.
\section{\uppercase{Motivation}}
\cite{Yanxi} first addressed the de-fencing problem via inpainting of fence occlusions. \cite{Minwo} used multiple images for de-fencing, which significantly improves the performance due to availability of hidden information in additional frames. They used a deformable lattice detection method proposed in \cite{Minwoo} for fence detction. However, this is not a robust approach and fails for many real world images as shown in Fig. 1(c), 3(b), 4(b), 5(b). \cite{Vrushali} proposed an improved multi-frame de-fencing technique by using loopy belief propagation. However, there are two issues with their approach. Firstly, the work in \cite{Vrushali} assumed that motion between the frames is global. This assumption is invalid for more complex dynamic scenes where the motion is non global. Also, their method used an image matting technique proposed by \cite{Yuanjie} for fence detection which involves significant user interaction.
Recently, \cite{Yadong} proposed a soft fence detection method where visual parallax serves as the cue to distinguish fences from the unoccluded pixels. Therefore, in this position paper we explore these issues and propose techniques for automatic fence detection.

\begin{figure}[t]
\centering     %%% not \center
\subfigure[]{\label{fig:a}\includegraphics[width=24mm]{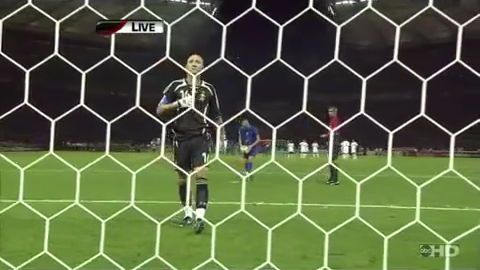}}
\subfigure[]{\label{fig:b}\includegraphics[width=24mm]{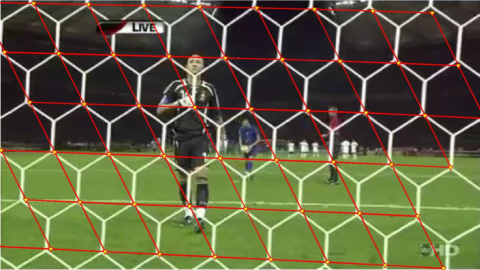}}
\subfigure[]{\label{fig:c}\includegraphics[width=24mm]{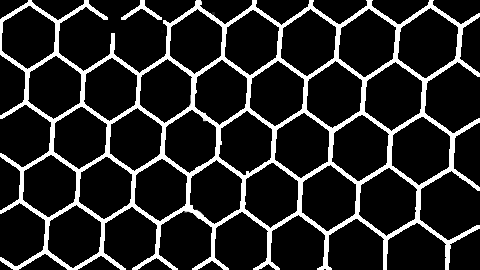}}
%S\subfigure[]{\label{fig:e}\includegraphics[width=40mm]{yanxi_Error_plot}}
\caption{Fence detection: (a) 1st frame from the video. (b) Output of the lattice detection algorithm \cite{Minwoo}. (c) Fence pixels detected using Gabor filter.}
\end{figure}
\begin{figure}[t]
\centering     %%% not \center
\subfigure[]{\label{fig:a}\includegraphics[width=24mm]{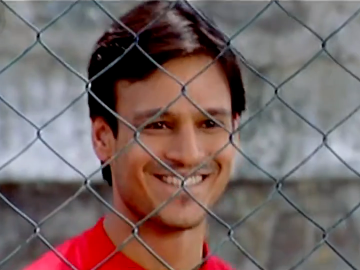}}
\subfigure[]{\label{fig:b}\includegraphics[width=24mm]{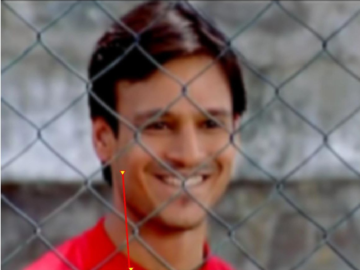}}
\subfigure[]{\label{fig:c}\includegraphics[width=24mm]{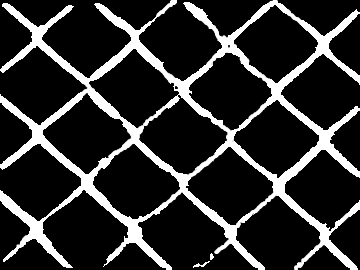}}
%S\subfigure[]{\label{fig:e}\includegraphics[width=40mm]{yanxi_Error_plot}}
\caption{Fence detection: (a) 1st frame from the video. (b) Output of the lattice detection algorithm \cite{Minwoo}. (c) Fence pixels detected using Gabor filter.}
\end{figure}
\begin{figure}[t]
\centering     %%% not \center
\subfigure[]{\label{fig:a}\includegraphics[width=36mm]{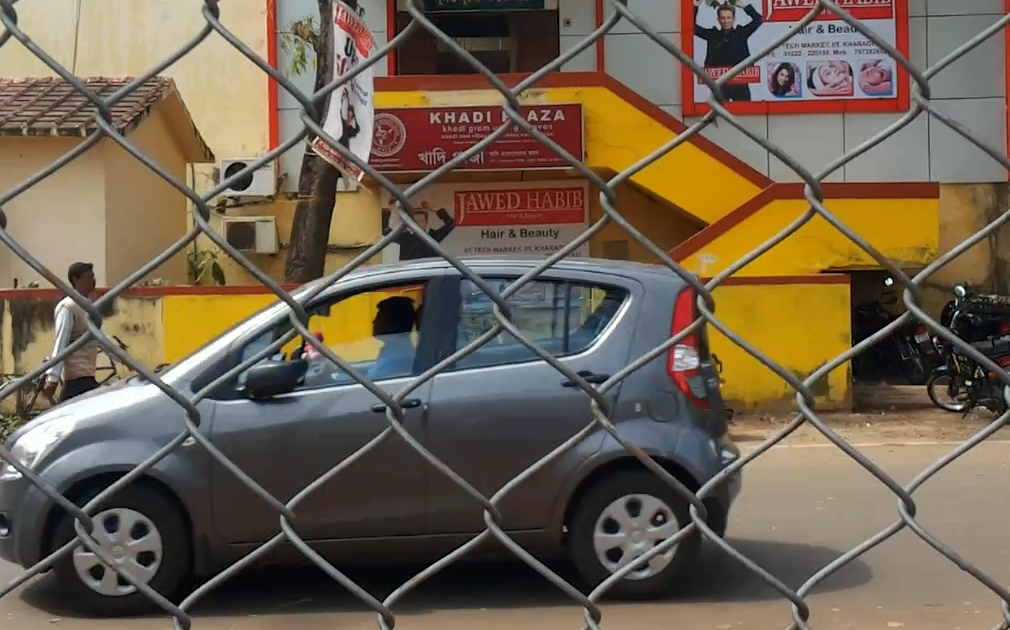}}
\subfigure[]{\label{fig:b}\includegraphics[width=36mm]{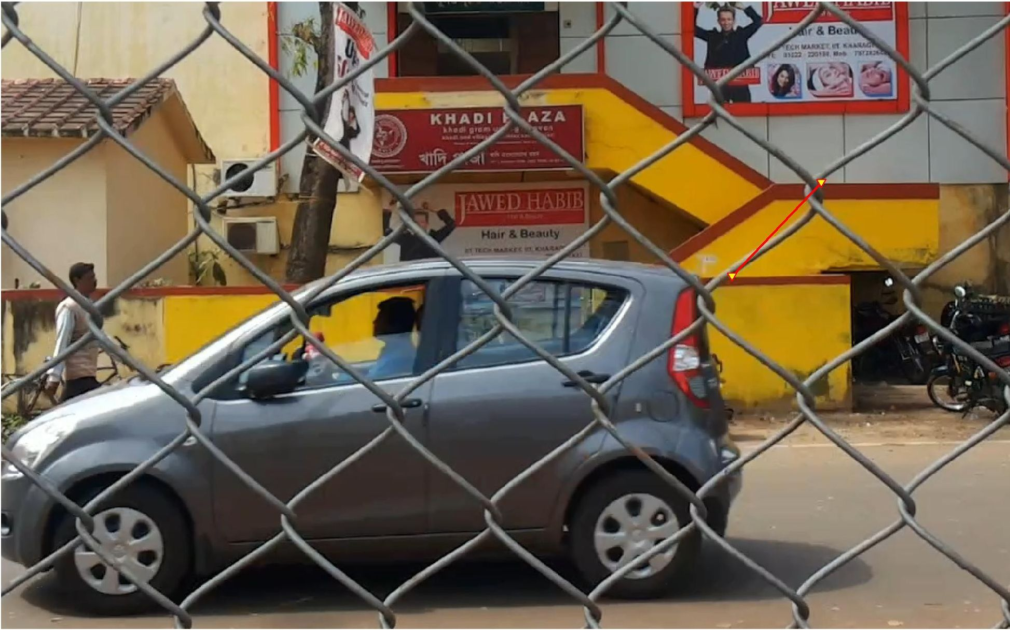}}
\subfigure[]{\label{fig:c}\includegraphics[width=36mm]{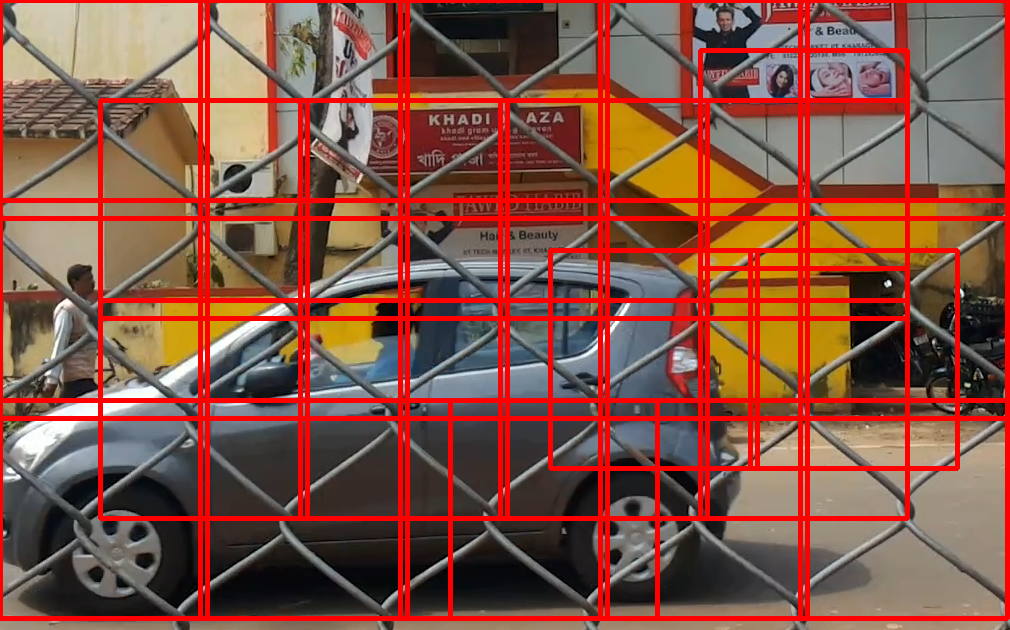}}
\subfigure[]{\label{fig:c}\includegraphics[width=36mm]{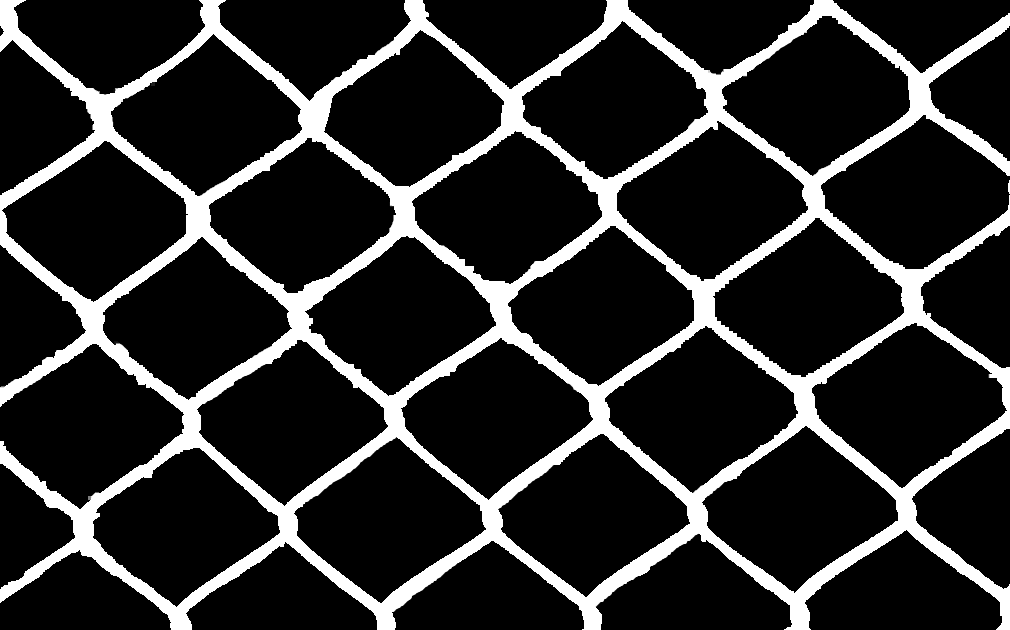}}
%S\subfigure[]{\label{fig:e}\includegraphics[width=40mm]{yanxi_Error_plot}}
\caption{Fence detection: (a) 1st frame from the video. (b) Output of the lattice detection algorithm \cite{Minwoo}. (c) Fence pixels detected using machine learning approach. (d) Fence mask.}
\end{figure}
 
The simplest way is to treat the fence detection as a segmentation problem. However for real world problems automated segmentation algorihms fail if the foreground and background layers are of similar color. We employ the graph-cuts based segmentation algorithm proposed by \cite{Boykov2001,Boykov2004,Kolmogorov2004}, on Fig. 1(a) using the Matlab wrapper by \cite{Bagon2006}. The segmentation result for the same is shown in Fig. 1(b). However, the automatic segmentation algorithm failed to detect the fence properly. Also, the method of \cite{Minwoo} failed to detect many fence pixels in Fig. 1(c). 

We propose two automatic approches to tackle fence detection. Since the fences have a strong directional property, we are motivated to employ a Gabor filter to detect them. To demonstrate the effectiveness of our proposed Gabor filter based technique and to compare with the state-of-the-art lattice detection method \cite{Minwoo} we used a real video from Youtube. We notice that the method of \cite{Minwoo} has wrongly detected fence pixels as shown in Fig. 3(b) and no pattern is detected in Fig, 4(b). Our proposed Gabor filter fence detection results are shown in Figs. 3(c), 4(c) wherein the fences have been properly detected. 

Secondly, we propose a machine learning based approach to the problem. We tested the technique on a frame from another video of real-world traffic shown in Fig. 5(a). It was found that the technique of \cite{Minwoo} failed to detect fence pixels shown in Fig. 5(b) but our proposed machine learning based approach detected the fence shown in Fig. 5(c). Note that the two methods mentioned above are completely automatic and require no user intervention.  

\section{\uppercase{Our Framework}}
We propose to use the following degradation model for the de-fencing problem
\begin{equation}
\textbf{y}_{m} = \textbf{O}_{m}\textbf{H}_{m}\textbf{W}_{m}\textbf{x} + \textbf{n}_{m}
\end{equation}
where $\textbf{y}_{m}$'s are the observations containing fences obtained from the captured video, $\textbf{x}$ is the de-fenced image, $\textbf{H}_{m}$ is blur operator for each frame, $\textbf{W}_{m}$ models the relative motion between frames, $\textbf{O}_{m}$ is obtained from the binary fence masks and $\textbf{n}_{m}$ is Gaussian noise.      

\subsection{Fence Detection}

\subsubsection{Gabor Filter approach}
Fences in general are inherently directional in nature. This property can be exploited by using directional filters. We employ the 2D Gabor filter proposed in \cite{Daugman} to our problem. It is given by, 
\begin{equation}
g(x,y;\lambda, \theta , \psi , \sigma ,\gamma) = \exp \big(-\frac{x^{\prime 2} + \gamma^{2} y^{\prime 2}}{2\sigma^{2}}\big) \cos\big(2\pi \frac{x^{\prime}}{\lambda}+\psi\big)
\end{equation}
Where $\lambda$ represents the wavelength, $\theta$ represents the orientation angle, $\psi$ represents the phase offset, $\sigma$ represents the standard deviation and $\gamma$ represents the aspect ratio. The parameter $\theta$ can be used to specify the orientation of the fences in the image. Here $\theta$ can be varied between 0 - 360 degrees based on the fence orientation. For example for fence detection, in Fig. 4(a) we use the Gabor filter with orientation angles 45, 225 degrees and other parameter values chosen as $\lambda=4$, $\psi=0$, $\gamma=0.5$, and $\sigma=4$. As shown in Fig. 4(c) the fence mask is detected accurately.

\subsubsection{Machine learning approach}
It is amply demonstrated in the literature that HOG features have been successful in many recognition and object classification problems. In this position paper, we propose a supervised learning approach to detect the fence pixels using HOG features \cite{Dalal}.

Firstly, all the dataset images are preprocessed by histogram normalization to reduce the effects of illumination changes. Each training image from dataset of ($100$ positives and $100$ negatives) is divided into non-overlapping cells of size $8 \times 8$ pixels and then the image gradient is computed in terms of magnitude as well as orientation. At every pixel in the \textit{cell}, the orientation is quantized into one of the nine bins, weighted based on its magnitude. The orientation bins are evenly spaced over $0-180$ degrees with each bin of size $20$ degrees. Finally, a histogram with the $9$ orientations is computed for each cell to form a feature vector of size $9\times 8\times 8$. A region of 4 \textit{cells} is clustered together to form a block and every neighboring block has an overlap of 2 \textit{cells}. A single block is thus represented by a feature vector of length $4\times 9 \times 8\times 8$. Every block which consists of un-normalized features from the \textit{cells}, is normalized by its $L2$ norm. Finally, all the feature vectors from the blocks are concatenated to obtain a single large feature vector of size 4752 corresponding to a single training image.

Since SVM classifiers were originally used for binary classification problems, we have chosen it for our problem. The extracted HOG features were used for training  an SVM for the classification of fence/non-fence. We used the RBF kernel which is given as $k(x_{i},x_{j}) = \exp(\gamma\parallel x_{i}-x_{j}\parallel^{2})$ where the parameter $\gamma$ and the misclassification penalty $C$ are found by a 5-fold cross validation.

As shown in Fig. 5(c), we use a sliding window to densely scan the test image from top to bottom and left to right at different scales. For each detector window, HOG features are extracted and fed to the trained SVM classifier to classify the sub-image as fence or non-fence. We replace positions of detected windows with a template binary mask to generate the final fence mask shown in Fig. 5(d).
\subsection{Motion Estimation}
The basic idea behind our method is that occluded image data in the reference frame is uncovered in other frames of the captured video. Motion estimation is to fuse the information uncovered in the other images for filling in occlusions in the reference frame. The relative shifts among the images have to be estimated in the degradation model of Eq. 1, to effect the image operations corresponding to $\textbf{W}_{m}$. Recently, \cite{Thomas} proposed an optical flow estimation technique, where they have integrated descriptor matching in a variational framework. This method is very effective in detecting sub pixel motion shifts in real world cases images without occlusions. However, for our application we need to accurately estimate optical flow for images with fences. When the optical flow for such images are estimated by \cite{Thomas}, we observe erroneous values around the fenced or occluded pixels. To avoid these errors, we smoothen observations using a Gaussian kernel, prior to using \cite{Thomas} to estimate the optical flow.
\subsection{Optimization}
We now formulate the optimization problem needed to solve the ill-posed inverse problem of image de-fencing. We minimize an objective function consisting of data fidelity term and a regularization term. We assume total variation (TV) of the de-fenced image as the regularization constraint. TV regularization is a well studied approach which preserves discontinuities in the reconstructed image \cite{Pascal,Konstantinos}.

The de-fenced image is the solution of the following optimization problem
\begin{equation}
\arg\min_{\textbf{x}} \frac{1}{2} \sum_{m=1}^{p}\parallel  \textbf{y}_{m}-\textbf{O}_{m}\textbf{H}_{m}\textbf{W}_{m}\textbf{x}\parallel _{2}^{2} +  \mu \parallel  \nabla  \textbf{x} \parallel_{1} 
\end{equation}
where $p$ is the number of frames chosen from the video and $\mu$ is the regularization parameter.
The above problem can also be written in a constrained framework as
\begin{equation}
\begin{split}
\arg\min_{\textbf{x}}  \frac{1}{2} \sum_{m=1}^{p}\parallel  \textbf{y}_{m}-\textbf{O}_{m}\textbf{H}_{m}\textbf{W}_{m}\textbf{x}\parallel _{2}^{2} +  \mu \parallel  \textbf{d} \parallel_{1} \\ s.t. \hspace{5pt}\textbf{d} = \nabla  \textbf{x}
\end{split}
\end{equation}
%$ \lim_{x \to \infty} \exp(-x) = 0 $
The above optimization framework is a combination of both $l1$ and $l2$ terms and hence difficult to solve. We employ the split Bregman iterative framework described in \cite{Tom} to solve the above problem. We use an alternative unconstrained formulation as
\begin{equation}
\begin{split}
 \arg\min_{\textbf{x}}  \frac{1}{2} \sum_{m=1}^{p}\parallel  \textbf{y}_{m}-\textbf{O}_{m}\textbf{H}_{m}\textbf{W}_{m}\textbf{x}\parallel _{2}^{2} \\ + \mu \parallel  \textbf{d} \parallel_{1} 
+\frac{ \lambda }{2}  \parallel \textbf{d}- \nabla \textbf{x} \parallel_{2}^{2}
\end{split}
\end{equation}
where $\lambda$ is the shrinkage parameter.
The iterates to solve the above equation are as
\begin{equation}
\begin{split}
 [\textbf{x}^{k+1},\textbf{d}^{k+1}] = \arg\min_{\textbf{x},\textbf{d}} \ \frac{1}{2} \sum_{m=1}^{p}\parallel  \textbf{y}_{m}-\textbf{O}_{m}\textbf{H}_{m}\textbf{W}_{m}\textbf{x}^{k}\parallel _{2}^{2} \\+ \mu \parallel  \textbf{d}^{k} \parallel_{1}
+\frac{ \lambda }{2}  \parallel \textbf{d}^{k}- \nabla \textbf{x}^{k} + \textbf{b}^{k}\parallel_{2}^{2}
\end{split}
\end{equation}
\begin{equation}
\textbf{b}^{k+1} = \nabla \textbf{x}^{k+1} + \textbf{b}^{k} -\textbf{d}^{k+1}
\end{equation}
We can now split the above problem into two sub-problems as \\
\\
\textbf{Sub Problem 1:}
\begin{equation}
\begin{split}
 [\textbf{x}^{k+1}] = \arg\min_{\textbf{x}} \ \frac{1}{2} \sum_{m=1}^{p}\parallel  \textbf{y}_{m}-\textbf{O}_{m}\textbf{H}_{m}\textbf{W}_{m}\textbf{x}^{k}\parallel _{2}^{2} \\ + \frac{ \lambda }{2}  \parallel \textbf{d}^{k}- \nabla \textbf{x}^{k} + \textbf{b}^{k}\parallel_{2}^{2}
\end{split}
\end{equation}
This sub-problem is solved by a gradient descent method.\\
\\
\textbf{Sub Problem 2:}
\begin{equation}
 [\textbf{d}^{k+1}] = \arg\min_{\textbf{d}} \mu \parallel  \textbf{d}^{k} \parallel_{1} + \frac{ \lambda }{2}  \parallel \textbf{d}^{k}- \nabla \textbf{x}^{k+1} + \textbf{b}^{k}\parallel_{2}^{2}
\end{equation}
The above sub-problem can be solved by applying the shrinkage operator as follows
\begin{equation}
\textbf{d}^{k+1}=shrink(\nabla \textbf{x}^{k+1}+\textbf{b}^{k},\frac{\lambda}{\mu})
\end{equation}
\begin{equation}
%\begin{split}
\textbf{d}^{k+1} = \frac{\nabla \textbf{x}^{k+1}+\textbf{b}^{k}}{\mid \nabla \textbf{x}^{k+1}+\textbf{b}^{k} \mid}*max(\mid \nabla \textbf{x}^{k+1}+\textbf{b}^{k} \mid - \frac{\lambda}{\mu},0) 
%\textbf{d}^{k+1}= \frac{\nabla \textbf{x}^{k+1}+\textbf{b}^{k}}{\mid \nabla \textbf{x}^{k+1}+\textbf{b}^{k} \mid}\ntimes max(\mid \nabla \\
%\textbf{x}^{k+1}+\textbf{b}^{k} \mid-\frac{\lambda}{\mu},0)
%where\hspace{4pt} shrink(\textbf{d},\lambda) = \frac{\textbf{d}}{ \mid \textbf{d} \mid}*max(\mid \textbf{d} \mid-\lambda,0)
%\end{split}
\end{equation}
The update for \textbf{b} is as
$ \textbf{b}^{k+1} = \nabla \textbf{x}^{k+1} + \textbf{b}^{k} - \textbf{d}^{k+1}$.
 We tune the parameters $\mu$, $\lambda$ to obtain the best estimate of the de-fenced image.

%\subsection{Algorithm}
%The following algorithm was used
%\begin{algorithm}
%\caption{Split Bregman De-fencing}
%\begin{algorithmic}
%\STATE $Input:\lambda, \mu$, $\textbf{y}_{m}$, $tol_{\textbf{x}}$, $tol_{\textbf{y}}$
%\STATE $Initialise:\textbf{d}^{0}=0, \textbf{b}^{0}=0$, $\textbf{y}_{m} = \textbf{y}^{old}_{m},$ 
%\STATE $\textbf{x}^{0}=random \hspace{3pt}numbers\hspace{3pt} generated\hspace{3pt} from\hspace{3pt} uniform\hspace{3pt} pdf$
%\WHILE{\(\parallel \textbf{y}_{m}^{k} - \textbf{y}_{m}^{k+1} \parallel _{2}^{2} > tol_{\textbf{y}}\)}
%\STATE $\textbf{y}^{k}_{m}=\textbf{y}^{old}_{m}$
%\WHILE{\(\parallel \textbf{x}^{k} - \textbf{x}^{k+1} \parallel _{2}^{2} > tol_{\textbf{x}}\)}
%\STATE $ \textbf{x}^{k+1}= \arg\min_{\textbf{x}} \ \frac{1}{2} \sum_1^n\parallel \textbf{y}_{m}^{k}-\textbf{O}_{m}\textbf{H}_{m}\textbf{W}_{m}\textbf{x}^{k}\parallel _{2}^{2} + 
%  \frac{ \lambda }{2}  \parallel \textbf{d}^{k}- \nabla \textbf{x}^{k} + \textbf{b}^{k}\parallel_{2}^{2} $
%\STATE $ \textbf{d}^{k+1}=shrink(\nabla \textbf{x}^{k+1}+\frac{\lambda}{\mu}\textbf{b}^{k})$
%\STATE $ \textbf{d}^{k+1} = \frac{\nabla \textbf{x}^{k+1}+\textbf{b}^{k}}{\mid \nabla \textbf{x}^{k+1}+\textbf{b}^{k} \mid}*max(\mid %\nabla \textbf{x}^{k+1}+\textbf{b}^{k} \mid - \frac{\lambda}{\mu},0)$
%\STATE $ \textbf{b}^{k+1}= \nabla \textbf{x}^{k+1} + \textbf{b}^{k} - \textbf{d}^{k+1}$
%\ENDWHILE
%\STATE $ \textbf{y}_{m}^{k+1}=\textbf{y}_{m}^{k}+\textbf{y}_{m}^{old}-\textbf{O}_{m}\textbf{H}_{m}\textbf{W}_{m}\textbf{x}^{k}$
%\ENDWHILE
%\end{algorithmic}
%\end{algorithm}

\section{\uppercase{Experimental Results}}

For both the synthetic and real-world cases we choose four images from the corresponding video sequence. Ideally, the images should be chosen in such a way that occluded information in the reference frame reappears in the adjacent frames. The de-fencing procedure is carried out individually in each color channel and the results combined to generate the RGB color image.

For synthetic experiments, we use the image of a tiger shown in Fig. 6. We shifted this image by (-8,- 8), (8, 8) and (15, 15) pixels to obtain four different frames. Simulating a fence of 7 pixel thickness, we removed image data from all 4 frames. The proposed algorithm was then applied with an initial estimate consisting of random numbers obtained from a uniform PDF. A value of $\lambda=0.01$ and $\mu=0.00001$ are used in the optimization method. 

The reconstructed image shown in Fig. 6(c) was found to have a PSNR of 39.8377 and SSIM of 0.9976. These quantitative results clearly validate the proposed algorithm. Also, the convergence of the proposed method can be seen in Fig. 6 (d) where we have plotted error vs number of Bregman iterations. The algorithm converges quickly during the first few iterations.
\begin{figure}[t]
\centering     %%% not \center
\subfigure[]{\label{fig:a}\includegraphics[width=37mm]{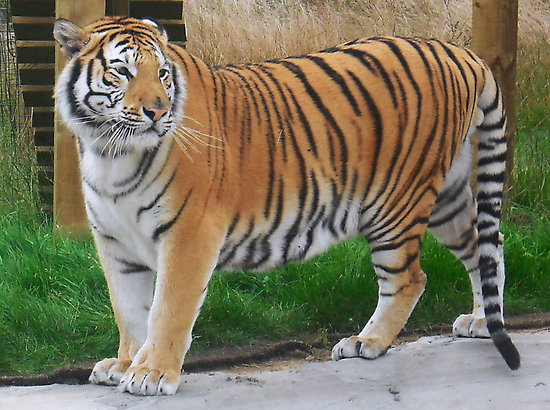}}
\subfigure[]{\label{fig:b}\includegraphics[width=37mm]{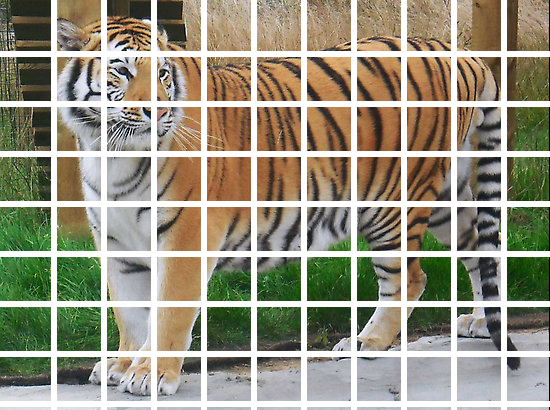}}
\subfigure[]{\label{fig:c}\includegraphics[width=37mm]{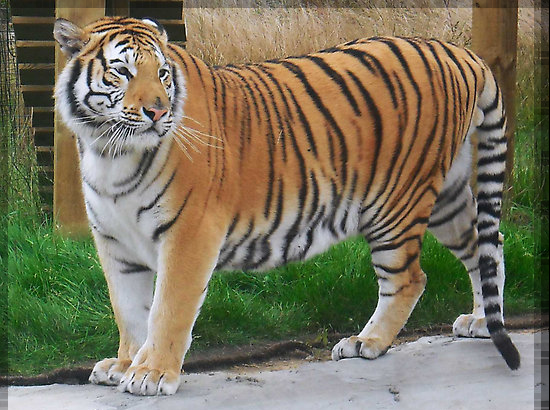}}
\subfigure[]{\label{fig:d}\includegraphics[width=37mm]{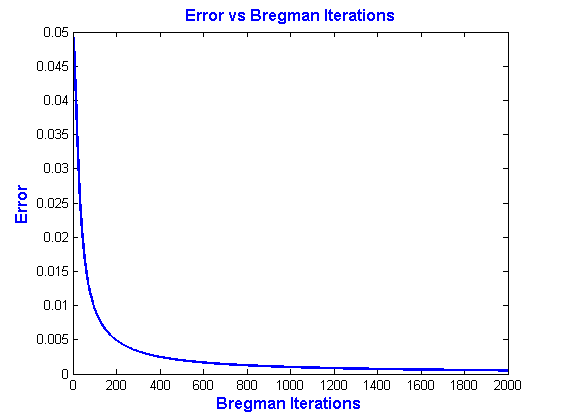}}
\caption{(a) Original image. (b) Fenced image. (c) De-fenced image estimated using the proposed algorithm. (d) Error analysis over bregman iterations.}
\end{figure}
Next, we have conducted experiments on a video from the 'Prison Break' TV sitcom obtained from Youtube. We have taken four frames for our algorithm, two of which are shown in Figs. 7(a), 7(b). We observed that the relative motion in the body region is noticeable whereas in the other parts is less. Therefore, inter-frame motion is non-global which makes the problem more challenging. We first computed the fence masks using the Gabor filter approach. Motion or optical flow between the frames were computed using the method proposed by \cite{Thomas}. Fig 7(c) shows the result of \cite{Minwo}. We observe many artifacts at the lips, shirt and hair of the person shown in the close-ups of Fig. 7(c). The proposed algortithm reconstructs the de-fenced image as shown in Fig. 7(d). We observe that the occlusions in the body region are completely filled-in with hardly any artifacts.
\begin{figure}[t]
\centering     %%% not \center
\subfigure[]{\label{fig:a}\includegraphics[width=32mm]{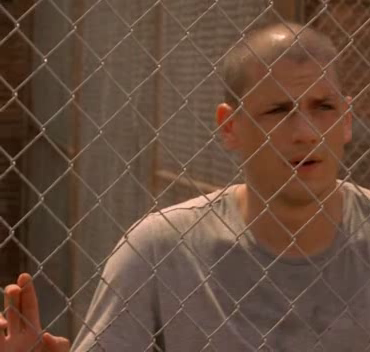}}
\subfigure[]{\label{fig:b}\includegraphics[width=32mm]{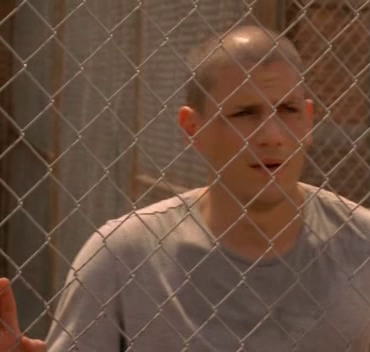}}
\subfigure[]{\label{fig:c}\includegraphics[width=32mm]{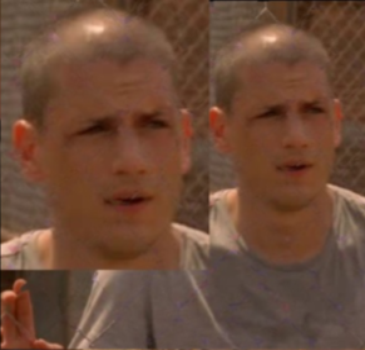}}
\subfigure[]{\label{fig:d}\includegraphics[width=32mm]{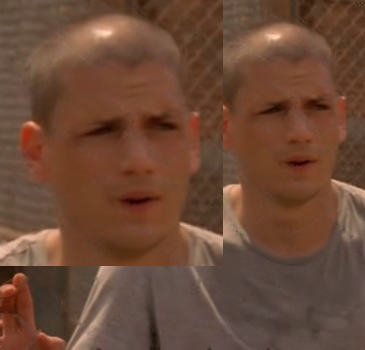}}
%S\subfigure[]{\label{fig:e}\includegraphics[width=40mm]{yanxi_Error_plot}}
\caption{(a), (b) 1st and 4th frames chosen from the video. (c) De-fenced image using \cite{Minwo}. (d) De-fenced image estimated using the proposed algorithm.}
\end{figure}

\begin{figure}[t]
\centering     %%% not \center
\subfigure[]{\label{fig:a}\includegraphics[width=37mm]{saathiya_obs1}}
\subfigure[]{\label{fig:b}\includegraphics[width=37mm]{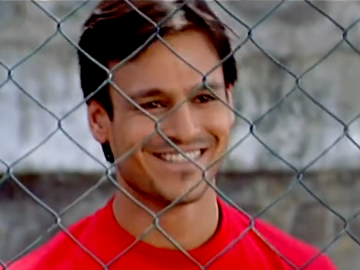}}
\subfigure[]{\label{fig:c}\includegraphics[width=37mm]{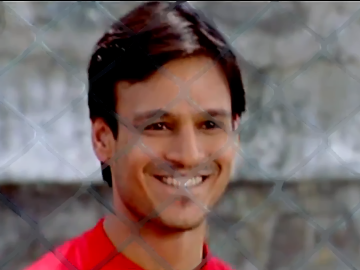}}
\subfigure[]{\label{fig:d}\includegraphics[width=37mm]{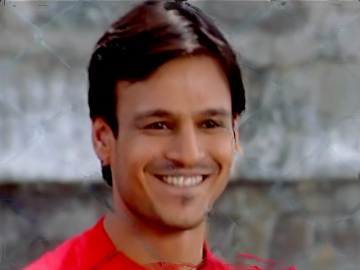}}
\caption{(a), (b) 1st and 4th frames chosen from the video. (c) Inpainting result of \cite{Konstantinos}. (d) De-fenced image estimated using the proposed algorithm.}
\end{figure}

Lastly, we move towards a more challenging problem, wherein we have used a video of a song from an Indian movie downloaded from Youtube. We have chosen four frames for our experimentation, two of them are shown in Figs. 8(a), (b). We notice a large amount of relative motion between the frames especially in the person's body and lesser amount of motion in the background.  We applied an inpainting technique proposed by \cite{Konstantinos} on the frame shown in Fig. 8(a) and the result is shown in Fig. 8(c). We noticed that fence pattern was still visible in the inpainted result particularly on the face portion. However, our multi-frame optimization framework uses actual data uncovered in the adjacent frames to effectively fill-in the missing information in the reference image as shown in Fig. 8(d).

We also show challenging cases where our automated fence detection algorithms fail. Fig. 9(b) shows the detected fence obtained using the proposed Gabor filter approach, we observe that some fence pixels are not detected due to similarity in color of both fence and car tyres. We show another example using our machine learning based approach in Fig. 9(d). We observe that the proposed approach failed to detect the fence pixels due to significant deformation in its shape. As a part of future work, we are investigating how to robustly detect fences when the camera is not fronto-parallel to the scene.

%\subsection{Comparisons with earlier work :}
%\subsection{Failure Cases :}

\section{\uppercase{Conclusions}}
\label{sec:conclusion}
In this position paper, we proposed an automatic image de-fencing algorithm for real-world videos. We divided the problem of image de-fencing into three tasks and proposed an automatic approach for each one of them. We formulated an optimization framework and solved for the inverse problem using the split Bregman technique assuming total variation of the de-fenced image as the regularization constraint. We have evaluated our proposed algorithm on both synthetic and real-world videos. The obtained results show the effectiveness of our proposed algorithm. As part of future work, we are investigating how to optimally choose the frames from the video.

\begin{figure}[t]
\centering     %%% not \center
\subfigure[]{\label{fig:a}\includegraphics[width=37mm]{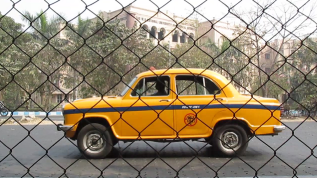}}
\subfigure[]{\label{fig:b}\includegraphics[width=37mm]{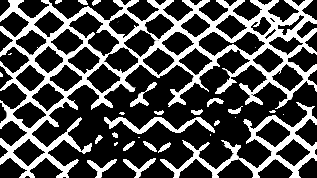}}
\subfigure[]{\label{fig:c}\includegraphics[width=37mm, height=27mm]{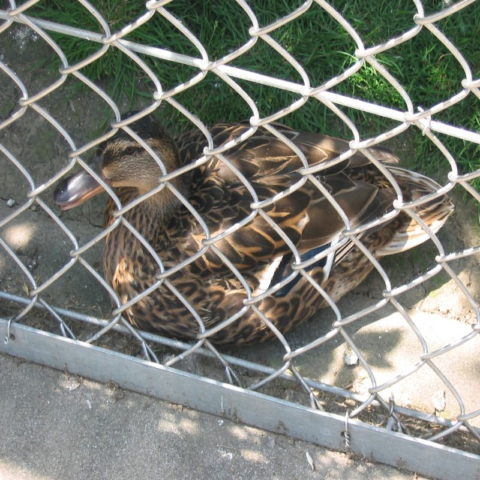}}
\subfigure[]{\label{fig:d}\includegraphics[width=37mm, height=27mm]{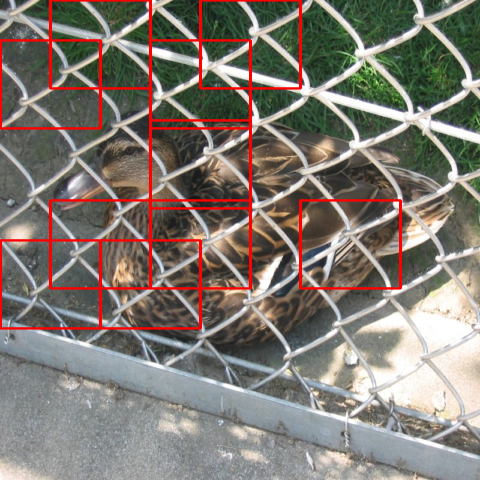}}
\caption{(a) 1st frame from the video. (b) Fence pixels detected using proposed Gabor filter. (c) 1st frame from the video. (d) Fence pixels detected using proposed machine learning approach.}
\end{figure}

%\section*{\uppercase{Acknowledgements}}
%\vfill
\bibliographystyle{apalike}
{\small
\bibliography{ganesh_db}

\begin{thebibliography}{}

\bibitem[Bagon, 2006]{Bagon2006}
Bagon, S. (2006).
\newblock http://www.wisdom.weizmann.ac.il/~bagon.

\bibitem[Bertalmio et~al., 2000]{Bertalmio}
Bertalmio, M., Sapiro, G., Caselles, V., and Ballester, C. (2000).
\newblock Image inpainting.
\newblock {\em ACM SIGGRAPH}, pages 417--424.

\bibitem[Boykov and Kolmogorov, 2004]{Boykov2004}
Boykov, Y. and Kolmogorov, V. (2004).
\newblock An experimental comparison of min-cut/max-flow algorithms for energy
  minimization in vision.
\newblock {\em IEEE Trans. Patt. Anal. Mach. Intell.}, 26(9):1124--1137.

\bibitem[Boykov et~al., 2001]{Boykov2001}
Boykov, Y., Veksler, O., and Zabih, R. (2001).
\newblock Efficient approximate energy minimization via graph cuts.
\newblock {\em IEEE Trans. Patt. Anal. Mach. Intell.}, 20(12):1222--1239.

\bibitem[Brox and Malik, 2011]{Thomas}
Brox, T. and Malik, J. (2011).
\newblock Large displacement optical flow: Descriptor matching in variational
  motion estimation.
\newblock {\em IEEE Trans. Patt. Anal. Mach. Intell.}, 33(3):500--513.

\bibitem[Criminisi et~al., 2004]{Criminisi_tip}
Criminisi, A., Perez, P., and Toyama, K. (2004).
\newblock Region filling and object removal by exemplar-based image inpainting.
\newblock {\em IEEE Trans. Image Proc.}, 13(9):1--13.

\bibitem[Dalal and Triggs, 2005]{Dalal}
Dalal, N. and Triggs, B. (2005).
\newblock Histograms of oriented gradients for human detection.
\newblock In {\em Proc. IEEE Conf. Computer Vision and Pattern Recognition},
  pages 1--8.

\bibitem[Daugman, 1985]{Daugman}
Daugman, J.~G. (1985).
\newblock Uncertainty relation for resolution in space, spatial frequency, and
  orientation optimized by two-dimensional visual cortical filters.
\newblock {\em Journal of the Optical Society of America}, 2(7):1160--1169.

\bibitem[Getreuer, 2012]{Pascal}
Getreuer, P. (2012).
\newblock Total variation inpainting using split \uppercase{B}regman.
\newblock {\em Image Processing On Line}, 2:147--157.

\bibitem[Goldstein and Osher, 2009]{Tom}
Goldstein, T. and Osher, S. (2009).
\newblock The split \uppercase{B}regman method for $l1$ regularized problems.
\newblock {\em SIAM Journal on Imaging Sciences}, 2(2):323--343.

\bibitem[Hays and Efros, 2007]{James}
Hays, J. and Efros, A.~A. (2007).
\newblock Scene completion using millions of photographs.
\newblock {\em ACM Transactions on Graphics}, 26(3):1--7.

\bibitem[Khasare et~al., 2013]{Vrushali}
Khasare, V.~S., Sahay, R.~R., and Kankanhalli, M.~S. (2013).
\newblock Seeing through the fence: Image de-fencing using a video sequence.
\newblock In {\em IEEE International Conference on Image Processing (ICIP)},
  pages 1351--1355.

\bibitem[Kolmogorov and Zabih, 2004]{Kolmogorov2004}
Kolmogorov, V. and Zabih, R. (2004).
\newblock What energy functions can be minimized via graph cuts?
\newblock {\em IEEE Trans. Patt. Anal. Mach. Intell.}, 26(2):147--159.

\bibitem[Liu et~al., 2008]{Yanxi}
Liu, Y., Belkina, T., Hays, J.~H., and Lublinerman, R. (2008).
\newblock Image de-fencing.
\newblock In {\em IEEE Conference on Computer Vision and Pattern Recognition},
  pages 1--8.

\bibitem[Mu et~al., 2014]{Yadong}
Mu, Y., Liu, W., and Yan, S. (2014).
\newblock Video de-fencing.
\newblock {\em IEEE Transactions on Circuit Systems and Video Technology},
  24(7):1111--1121.

\bibitem[Papafitsoros et~al., 2013]{Konstantinos}
Papafitsoros, K., Sch¨onlieb, C.-B., and Sengul, B. (2013).
\newblock Combined first and second order total variation inpainting using
  split \uppercase{B}regman.
\newblock {\em Image Processing On Line}, 3:112--136.

\bibitem[Park et~al., 2009]{Minwoo}
Park, M., Brocklehurst, K., Collins, R.~T., and Liu, Y. (2009).
\newblock Deformed lattice detection in real-world images using mean-shift
  belief propagation.
\newblock {\em IEEE Trans. Patt. Anal. Mach. Intell.}, 31(10):1804--1816.

\bibitem[Park et~al., 2010]{Minwo}
Park, M., Brocklehurst, K., Collins, R.~T., and Liu, Y. (2010).
\newblock Image de-fencing revisited.
\newblock In {\em Asian Conference on Computer vision}, pages 422--434.

\bibitem[Xu and Sun, 2010]{Xu}
Xu, Z. and Sun, J. (2010).
\newblock Image inpainting by patch propagation using patch sparsity.
\newblock {\em IEEE Trans. Image Proc.}, 19(5):1153--1165.

\bibitem[Zheng and Kambhamettu, 2009]{Yuanjie}
Zheng, Y. and Kambhamettu, C. (2009).
\newblock Learning based digital matting.
\newblock In {\em IEEE International Conference on Computer Vision}, pages
  889--896.

\end{thebibliography}
\vfill
\end{document}